\documentclass{Interspeech}

\interspeechcameraready

\title{What do self-supervised speech models know about Dutch? \\ Analyzing advantages of language-specific pre-training}

\author[affiliation={1}]{\qquad\qquad Marianne}{de Heer Kloots}
\author[affiliation={2}]{Hosein}{Mohebbi}
\author[affiliation={1}]{Charlotte}{Pouw}
\author[affiliation={2}]{\\Gaofei}{Shen}
\author[affiliation={1}]{Willem}{Zuidema}
\author[affiliation={3}]{Martijn}{Bentum}

\affiliation{Institute for Logic, Language and Computation}{University of Amsterdam}{The Netherlands}
\affiliation{Cognitive Science and Artificial Intelligence}{Tilburg University}{The Netherlands}
\affiliation{Centre for Language Studies}{Radboud University}{Netherlands}
\email{m.l.s.deheerkloots@uva.nl, H.Mohebbi@tilburguniversity.edu, c.m.pouw@uva.nl, G.Shen@tilburguniversity.edu, w.h.zuidema@uva.nl, martijn.bentum@ru.nl}

\usepackage{comment}

\begin{document}
\maketitle

\begin{abstract}
How language-specific are speech representations learned by self-supervised models? Existing work has shown that a range of linguistic features can be successfully decoded from end-to-end models trained only on speech recordings. However, it's less clear to what extent pre-training on specific languages improves language-specific linguistic information. Here we test the encoding of Dutch phonetic and lexical information in internal representations of self-supervised Wav2Vec2 models. Pre-training exclusively on Dutch improves the representation of Dutch linguistic features as compared to pre-training on similar amounts of English or larger amounts of multilingual data. This language-specific advantage is well-detected by trained clustering or classification probes, and partially observable using zero-shot metrics. Furthermore, the language-specific benefit on linguistic feature encoding aligns with downstream performance on Automatic Speech Recognition.
\end{abstract}

\section{Introduction}
In recent years, self-supervised learning (SSL) algorithms have been demonstrated to learn powerful representations of spoken language, in terms of both their downstream task performance and the richness of their embedding spaces. Despite being trained only on unlabeled speech recordings, neural SSL models vastly outperform acoustic baselines at encoding various levels of linguistic structure, including phonological \cite{tenboschPhonemicCompetitionEndend2023, bentumProcessingStressEndEnd2024}, syllabic \cite{deheerklootsHumanlikeLinguisticBiases2024} and lexical \cite{pasadWhatSelfSupervisedSpeech2024} features. Understanding the linguistic information these models encode is relevant not only for engineers interpreting model behaviour, but also for cognitive scientists studying language acquisition from spoken input \cite{lavechinSimulatingEarlyPhonetic2025, dupouxCognitiveScienceEra2018a}. 

Despite the wide interest for interpreting SSL model functioning, research has so far primarily focused on English linguistic features encoded by English-only or multilingually pre-trained models, with some notable exceptions \cite{shenEncodingLexicalTone2024, delafuenteLayerwiseAnalysisMandarin2024, dugonjicWhatHasLeBenchmark2024, milletSelfsupervisedSpeechModels2022a}. As a result, it remains somewhat unclear whether measures detecting linguistic structures in SSL models rely on truly language-specific representations or more language-general acoustic correlates. For downstream automatic speech recognition (ASR), both multi- and monolingually trained Wav2Vec2 \cite{DBLP:journals/corr/abs-2006-11477} models are known to generalize well even to languages not present in their training data \cite{riviereUnsupervisedPretrainingTransfers2020, babu22-XLS-R-Interspeech}. This could indicate that what these models learn about speech is not typically language-specific \cite{dieck22_interspeech}, in line with findings that internal activations of Wav2Vec2 models do not seem strongly influenced by the model's training language in distinguishing native vs. non-native phoneme contrasts \cite{milletSelfsupervisedSpeechModels2022a}. However, the same models do show distinct patterns in predicting brain activity for their pre-training language vs. a different language \cite{milletRealisticModelSpeech2022a}, and crosslingual comparisons of Wav2Vec2 models trained on tonal vs. non-tonal languages have found language-specific advantages in the representation of suprasegmental features \cite{shenEncodingLexicalTone2024, delafuenteLayerwiseAnalysisMandarin2024}.

A key challenge in distinguishing language-specific vs. language-general speech processing by neural SSL models is the variability in measures and datasets used to identify language-specific representations. While some studies use `zero-shot' embedding space distance metrics to measure the saliency of native vs. non-native phoneme contrasts \cite{milletSelfsupervisedSpeechModels2022a}, other studies use trained classifiers to detect how well language-specific information can be decoded from internal representations \cite{shenEncodingLexicalTone2024, delafuenteLayerwiseAnalysisMandarin2024}. Hence, the differences in reported results could be driven by differences in the methodologies used for testing the models' linguistic capabilities.

\begin{figure}[t]
  \centering
  \includegraphics[width=\linewidth]{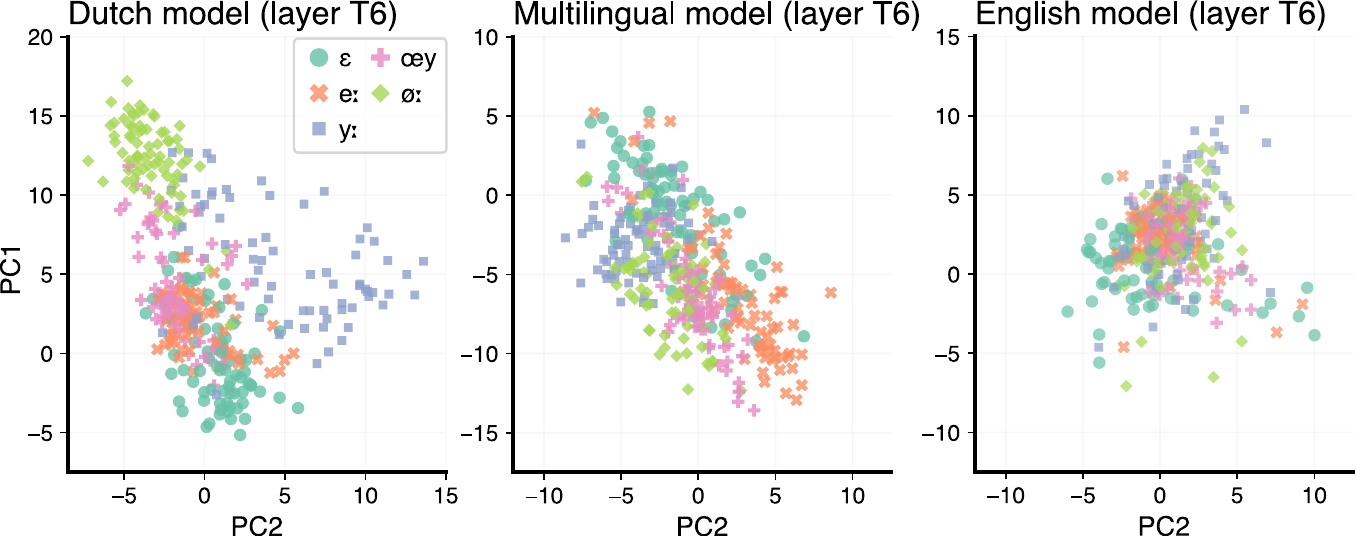}
  \caption{A set of high front vowels (phonemically distinctive in Dutch) is most discriminably represented in a model trained exclusively on Dutch. Data points are vowel occurrences sampled across 3 speakers in Multilingual LibriSpeech.}
  \label{fig:pca}
  \vspace{-3mm}
\end{figure}

Here, we aim to investigate language-specific representations of Dutch linguistic information in self-supervised speech models, as well as the effects of different analysis pipelines on the detection of such information. For this purpose, we curated the \emph{SSL-NL} evaluation set of publicly available Dutch speech data and alignments\footnote{We release the \emph{SSL-NL} dataset at \url{http://doi.org/10.5281/zenodo.15548947}}. We designed the \emph{SSL-NL} set to test the encoding of Dutch phonetic and lexical features in SSL speech representations, while allowing for comparisons across different analysis methods. We also pre-train a Wav2Vec2 model on the Dutch language exclusively\footnote{We release \emph{Wav2Vec2-NL} along with its training manifest and configuration at \url{http://doi.org/10.5281/zenodo.15550628}} as an additional resource for studying Dutch-specific speech representations. 
Comparing the Dutch model to a monolingual English and a multilingual model, we find that both phone- and word-level features are best encoded in the Dutch model (e.g. Figure \ref{fig:pca}), which is also reflected in its downstream performance on Automatic Speech Recognition. Notably, the detection of language-specific benefits varies across analysis measures and datasets, underscoring the critical role of data selection and analysis methods in assessing the impact of language-specific pre-training on self-supervised speech representations.

\section{Models}
To investigate the effect of language-specific pre-training on linguistic feature encoding, we compare three Wav2Vec2 models with identical architectures (7 CNN + 12 Transformer layers), but with varying amounts of Dutch and other languages in their pre-training data. 

We train \textbf{w2v2-nl} (\texttt{Wav2Vec2-NL}) on 831 hours of spoken Dutch, combining data from the Spoken Dutch Corpus (CGN; \cite{schuurmanCGNAnnotatedCorpus2003}), Multilingual Librispeech (MLS; \cite{pratapMLSLargeScaleMultilingual2020}), and CommonVoice (CV; \cite{ardilaCommonVoiceMassivelyMultilingual2020}). We sample 537 hours from CGN, including both Dutch and Flemish recordings but excluding telephone conversations (recordings with low sample rate) and sermons (poor recording quality). The remaining data includes recordings of spontaneous conversations and interviews, as well as read speech and news reports. We segment the CGN recordings into phrases, and use segments between 2 and 15 seconds in length as inputs for model training. From MLS, we sample 211 hours of the provided audiobook segments. The remaining 83 hours are sampled from CommonVoice, consisting of short segments of read aloud speech. Across the full training set, audio samples range between 2 and 20 seconds in length. We follow the original configuration for training Wav2Vec2 in \cite{baevskiWav2vecFrameworkSelfSupervised2020}, only modifying it to allow longer utterance length and per-device batch size. We use the fairseq toolkit \cite{ottFairseqFastExtensible2019} to pre-train w2v2-nl for 100k training steps using 4 Nvidia A100-40GB GPUs. 

In addition to our Dutch model, we include two additional models trained on other languages, as well as a model trained on non-speech acoustics: \textbf{fb-en}, the self-supervised base model from the original Wav2Vec2 release, pre-trained on 960 hours of English audiobook recordings from LibriSpeech \cite{baevskiWav2vecFrameworkSelfSupervised2020}; \textbf{fb-voxp-100k}, a multilingual base model pre-trained on 100k hours of European parliament recordings in 23 languages, including 4,5k hours of Dutch \cite{wangVoxPopuliLargeScaleMultilingual2021}; and a \textbf{nonspeech} base model trained on 600 hours of acoustic scenes from AudioSet \cite{milletSelfsupervisedSpeechModels2022a}\footnote{HuggingFace \cite{wolfHuggingFacesTransformersStateoftheart2020} identifiers:  \texttt{amsterdamNLP/Wav2Vec2-NL} (w2v2-nl); \texttt{facebook/wav2vec2-base} (fb-en); \texttt{facebook/\\wav2vec2-base-100k-voxpopuli} (fb-voxp-100k);\\ \texttt{ewandunbar/humanlike-speech-2022} (nonspeech)}.

\section{Representational analysis methods}
We create the \emph{SSL-NL} evaluation set by sampling speech recordings from two different datasets of spoken Dutch. We use a separate subset of MLS audiobook segments (held-out from w2v2-nl training data), as well as the IFADV corpus \cite{vansonIFADVCorpusFree2008} of face-to-face conversational speech. Based on the speech recordings and orthographic transcription pairs in each corpus, we obtain phone- and word-level forced alignments for both IFADV and MLS using the WebMAUS API. Phone- and word-level representations are extracted from each model layer by passing the full MLS audiobook segment or IFADV conversational turn as input to the model, and mean-pooling across frame representations within each phone or word respectively.

For our \emph{phonetic analyses} we collect samples of 37 Dutch phone categories, including 13 vowels (\textipa{a:, A, E, e:, \o:, I, i:, y:, 0, u:, o:, O, @}), 3 diphthongs (\textipa{Ei, \oe y, Au}) and 21 consonants (\textipa{p, b, t, d, k, g, V, f, v, s, z, S, x, G, h, m, n, N, r, l, j}). From MLS, we collect 6,086 phone occurrences in total from 11 different speakers; from IFADV we collect 3,294 phone occurrences in total from 9 speakers. We sample 15 occurrences from MLS and 10 occurrences from IFADV per speaker per phone, except for [\textipa{\o:}], which occurs more rarely. For our \emph{lexical analyses} we sample a total of 1,338 word occurrences (307 unique words) from MLS and 734 word occurrences (124 unique words) from IFADV, from which we select smaller subsets for our clustering and RSA analyses, as explained below.

\begin{figure*}[ht!]
  \centering
  \includegraphics[width=\textwidth]{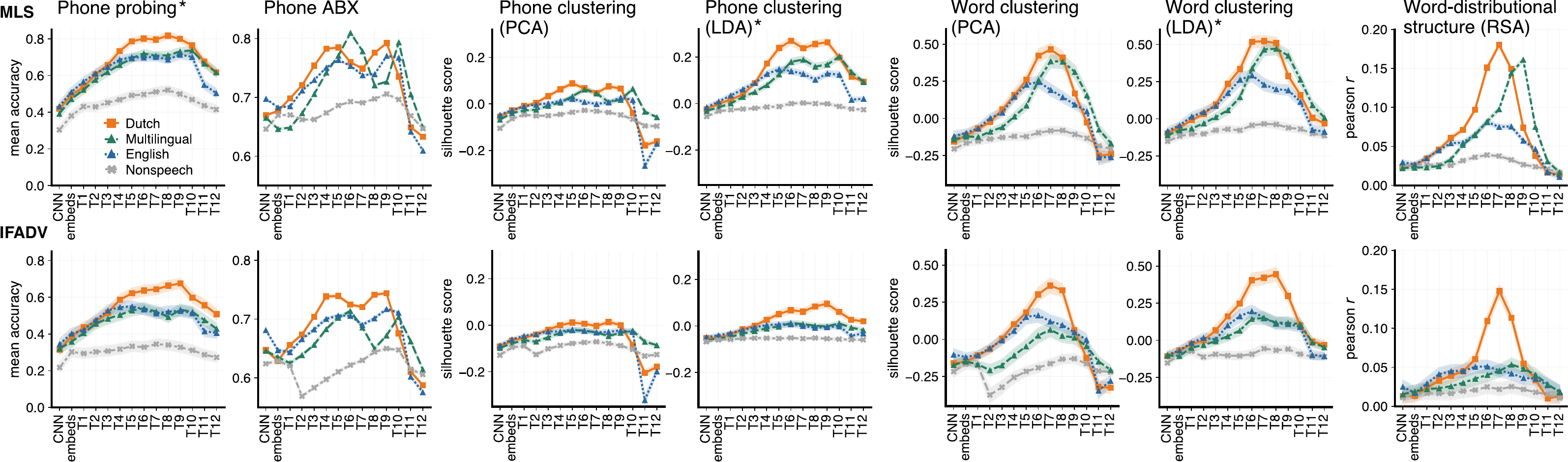}
  \caption{Layerwise phonetic and lexical analyses, across a read speech (MLS, top row) and a dialogue (IFADV, bottom row) dataset of spoken Dutch. Measures marked * involve optimized linear transforms, whereas others are computed zero-shot; shading indicates 95\% confidence intervals. The Dutch \textbf{w2v2-nl} model achieves highest scores across most analyses of Dutch phone and word encoding, though the size of this language-specific advantage varies considerably across analyses.}
  \label{fig:measure-comparison}
  \vspace{-2mm}
\end{figure*}

\subsection{Phonetic analyses}

\subsubsection{Phone identity probing}
Following previous studies on English (e.g. \cite{cormacenglishDomainInformedProbingWav2vec2022}), we first analyze the representation of Dutch phones in our models by fitting a linear transform to decode phone identity from model hidden layer representations. We fit separate multinomial logistic regression probes for each hidden layer on data from 8 MLS speakers and 6 IFADV speakers. We evaluate probe prediction accuracy on the remaining 3 speakers for each dataset. Chance accuracy for the phone identity probes is $\frac{1}{37}$, since we defined 37 phone categories.

\subsubsection{Phone ABX}
As an alternative to fitting trained classification probes, the encoding of phonetic categories in SSL models has also been investigated by directly measuring distances in model embedding spaces (e.g. \cite{milletSelfsupervisedSpeechModels2022a, poliModelingInitialState2024}). In the phone ABX task, we sample phone occurrence triplets A, B, and X, where A and X belong to the same phoneme category but B belongs to a different category. Phone representations are then evaluated on whether cosine similarity is greater for within-category embedding pairs (comparing A and X) than for between-category embedding pairs (comparing A and B). We construct phone ABX triplets based on a set of 59 unique Dutch phoneme contrasts (e.g. \textipa{/o:/} vs. /\textipa{O}/), while sampling A, B and X from different contexts and speakers. We evaluate phone ABX accuracy across a set of over 222K ABX triplets (3.7K per contrast) sampled from MLS, and over 66K triplets from IFADV (1.1K per contrast). Chance accuracy for the ABX measure is $\frac{1}{2}$, since accuracy for each ABX triplet is binary (testing if AX-similarity $>$ AB-similarity).

\subsubsection{Phone clustering}
Our final phone-level analysis method examines how well model embedding spaces cluster phone samples together according to our pre-defined Dutch phone categories. The degree of clustering can be measured using silhouette scores \cite{rousseeuwSilhouettesGraphicalAid1987, scikit-learn}, computed by comparing the mean distance between same-cluster samples with their mean distance to the next nearest cluster. We compute layerwise average silhouette scores across phone clusters after reducing model embedding dimensionality to 36 dimensions, either unsupervised (by applying Principal Component Analysis; PCA), or by optimizing a linear transform for separability between phone categories (by applying Linear Discriminant Analysis; LDA).
We evaluate the dimensionality reduced projections on a held-out set of speakers, after fitting them on a train set (applying the same split used for phone identity probing). Silhouette scores range between -1 and 1, with positive values indicating better cluster separability.

\subsection{Lexical analyses}
\subsubsection{Word clustering}
For word clustering analyses, we chose the 50 most frequent words from the BAK list of basic Dutch preschooler vocabulary \cite{mulder2009handreiking} within each evaluation subset. We sample up to 9 occurrences for every word, each from a different speaker and recording (448 occurrences in total for IFADV; 450 for MLS). We use two thirds of this set (6 out of 9 speakers) to compute and fit PCA and LDA transformations (reducing the embedding dimensionality to 49), and we use the final third for computing silhouette scores to evaluate the degree of clustering by word type.

\subsubsection{Word-distributional structure (Fasttext RSA)}
To analyze the encoding of Dutch word-distributional structure in our set of speech models, we evaluate representational similarity \cite{kriegeskorteRepresentationalSimilarityAnalysis2008a} of speech-based word embeddings to static text-based word embeddings from a Dutch Fasttext \cite{bojanowskiEnrichingWordVectors2017} model\footnote{\texttt{facebook/fasttext-nl-vectors}}. We compute representational similarity across a set of commonly occurring words from the same basic preschooler vocabulary list, sampling up to 3 occurrences from different speakers for every word (297 word types, 889 word tokens from MLS; 116 word types, 341 word tokens from IFADV). We measure representational similarity as the Pearson correlation between pairwise cosine distances for each of these word tokens and the corresponding distances in the text-based Fasttext word embeddings. In computing pairwise distances, we exclude word token pairs of the same type, to avoid effects of word identity.

\section{Results of representational analyses}

\subsection{Phonetic and lexical representations improve with language-specific pre-training}
Across most of our phonetic and lexical analyses, we observe a moderate to substantial advantage of the w2v2-nl model over its English and multilingual counterparts (Figure \ref{fig:measure-comparison}). This suggests that Dutch-specific pre-training enhances the encoding of Dutch phonetic and lexical features in self-supervised speech models. In the phonetic analyses, the small benefit reflects the fact that Dutch is phonetically relatively similar to English, as well as other European languages in the multilingual model’s pre-training data — many of the phone categories in our analysis set are shared by these languages. However, when zooming in on phones more specific to Dutch, such as the high-font vowels \textipa{[y:], [\o:]} and diphthong \textipa{[\oe y]}, we see that they are more distinctively represented along the first principal components of the Dutch model’s hidden layer representations, as compared to the multilingual and English models (Figure \ref{fig:pca}).

The English model also shows significantly higher scores than our non-speech baseline model on the lexical analyses. The majority of words in our lexical analysis sets are not particularly similar to English words, with some exceptions (e.g. \textipa{[bEt]}, \textipa{[hArt]}). Rather than recognizing Dutch word forms, it is more likely that the English model better encodes acoustic differences between words than the nonspeech model. Nevertheless, language-specific information does seem to help the encoding of lexical structure, as reflected by the higher scores of the Dutch and multilingual model (which included Dutch in its pre-training data) over the English one.

\subsection{Differences across analysis measures and datasets}
While SSL representations have been analyzed for linguistic structure before, many such analyses have been performed on read speech datasets. Moreover, studies differ in the type of analysis method applied, for example using either zero-shot ABX tasks or trained probing classifiers. We observe variability across both the different analysis measures and the different datasets we performed our analyses on (Figure \ref{fig:measure-comparison}). Across measures which make use of linear projections to optimize phone identity decoding (probing, LDA), Dutch phonetic encoding is clearly enhanced in the Dutch model compared to the other models. However, this is less clear in measures analyzing phonemic contrasts by directly comparing distances within each model’s representation space (ABX) or by analyzing the degree of phone clustering along a subset of its most variance-explaining components (PCA). The contrast between these results suggests that language-specific phonetic information may be encoded in a small subspace of model internal representations, which is decodable after a linear transformation, but not very prominently featured across the full embedding space. 

In contrast, our lexical analyses show that the word-level benefit of language-specific pre-training is observable using both zero-shot (PCA, RSA) and optimized (LDA) analysis measures. While phone clustering silhouette scores significantly improve after fitting LDA projections, word-level silhouette scores are relatively similar between PCA and LDA clustering metrics. Possibly, learned word identities are quite saliently represented in model embedding space, especially when mean-pooling across all 20 ms frame representations within words. Work on word representations in English SSL models has shown that word identity can be decoded from both mean-pooled and individual frame representations \cite{pasadWhatSelfSupervisedSpeech2024}, suggesting that this is not an effect of the mean-pooling operation itself.

\section{Downstream ASR performance}
We fine-tune our SSL models for speech-to-text transcription to examine whether the language-specific advantages in self-supervised representations also lead to improved performance on downstream ASR tasks. Each SSL model is fine-tuned on Dutch read-aloud speech from the CGN (component o), using 78 hours of training data while reserving 10 hours each for development and testing. ASR performance is evaluated using word error rate (WER) on the held-out CGN-o test set. Additionally, we assess performance on test sets from CV and MLS (both containing read-aloud speech), IFADV (dialogue speech), and N-Best \cite{kessens2007n}, a Dutch ASR benchmark featuring conversational telephone speech and news broadcast recordings. While we have no expectation of reaching state-of-the-art transcription performance with our models and fine-tuning set-up, we are interested in comparing relative model performance at this task to the observed differences in representational quality.

Across all test sets, the Dutch pre-trained \textbf{w2v2-nl} model consistently achieves lower WER than the English \textbf{fb-en} and multilingual \textbf{fb-voxp-100k} models (see Table \ref{tab:wer_results}). This suggests that the advantages of language-specific pre-training extend beyond self-supervised representations to downstream ASR performance. The performance of the Dutch model is followed by the multilingual and English models, aligning with the patterns observed in our phonetic and lexical analyses.

\begin{table}[h]
    \centering
    \begin{tabular}{lccccc}
        \hline
        Models         & CGN-o  & IFADV  & MLS  & CV  & N-Best \\
        \hline
        Dutch  & 10.4  & 65.6  & 15.4  & 21.0  & 25.2 \\
        English  & 21.5  & 84.4  & 32.3  & 47.9  & 53.1 \\
        Multilingual & 12.7  & 78.8  & 23.1  & 34.2  & 37.2 \\
        Nonspeech    & 43.5    & 94.9  & 58.6  & 78.4  & 75.7 \\
        \hline
    \end{tabular}
    \caption{WER results for models fine-tuned on the read-aloud component of the CGN and evaluated on held out test set (CGN-o), the test sets of two read-aloud corpora (MLS and CV), a dialogue corpus (IFADV) and a Dutch benchmark (N-Best).}
    \label{tab:wer_results}
    \vspace{-6mm}
\end{table}

\section{Discussion \& Conclusions}
We introduced the \emph{SSL-NL} evaluation set, and used it to compare a new monolingual Dutch Wav2Vec2 model against existing English and multilingual models. What do these models represent about the phonetic and lexical structure of Dutch? 

We find that linguistic information at both levels can be accurately decoded from each model's hidden layer representations, using a variety of methods. While accuracy is similar across models at the output of the CNN feature encoder, a language-specific advantage for the Dutch model emerges across the model's Transformer module. Thus, language-specific pre-training does benefit the encoding of phonetic and lexical structure in Wav2Vec2 hidden layer representations.

We observe interesting differences across analysis methods. Phone ABX tasks are widely used to evaluate self-supervised speech representations, including in studies of native- vs. non-native speech processing \cite{milletSelfsupervisedSpeechModels2022a, poliModelingInitialState2024}. However, we find that they may be a less sensitive measure for detecting language-specific information that is decodable by training classification or clustering probes on model representations. This is an important consideration for future work aiming to study the language-specificity vs. -generality of self-supervised representations — especially in models with relatively high representational dimensionalities. For example, large multilingually pre-trained Wav2Vec2 models may make use of language-specific subspaces to achieve similar linguistic encoding accuracy for individual languages as monolingually pre-trained models \cite{abdullah23_interspeech, babu22-XLS-R-Interspeech}.

Between the Dutch \textbf{w2v2-nl} model and the other models, we generally observe larger differences on the IFADV dialogue dataset than on the MLS read speech dataset. We believe this reflects an effect of the pre-training data domain beyond its language-specificity: while the Dutch model’s pre-training data included conversational speech, the English and multilingual models were respectively trained on read-aloud books and less spontaneous speech styles (parliament debates). The effect is particularly prominent for word-distributional structure, as measured by representational similarity between the speech models and the text-based Fasttext word embedding model. Word-distributional patterns are known to significantly differ between spoken and written language corpora \cite{bentum2019speech}, potentially limiting the generalizability of models trained on read speech \cite{dingemanseTextTalkHarnessing2022}. Our analyses show the benefit of training on conversational speech not only for enhancing the representation of conversation-level structures, but also for the encoding of smaller linguistic units such as phones and words.

When fine-tuning our set of SSL models on ASR, we find that the benefit of language-specific pre-training is also reflected in transcription accuracy. However, probe performance and downstream task accuracy are not necessarily directly related, and rankings of self-supervised models have shown stark differences when models are used as feature extractors (i.e. probed) vs. when they are fine-tuned for the evaluation task \cite{zaiem23b_interspeech, parcolletLeBenchmark20Standardized2024}. An interesting direction of future research could explore how the representation of linguistic features causally affects downstream text-transcription performance, for example using feature removal techniques to manipulate model representation spaces, as has been explored for the manipulation of gender representation in ASR models \cite{krishnan24_interspeech}.

In this study, we compared models trained on languages with relatively high phonetic similarity, but still found observable benefits of language-specific pre-training on encoding Dutch phone categories, and stronger effects on linguistically more complex features such as word-distributional structure. We note that we have not tested for specific kinds of lexical knowledge such as syntactic or semantic categories — to the extent that self-supervised models represent such higher-level information, we would expect stronger language-specific effects, but this remains to be tested. We hope that the w2v2-nl model and the SSL-NL evaluation set provide valuable resources for such further investigations into language-specific representations in self-supervised speech models.

\section{Acknowledgements}
We would like to thank David van Leeuwen and Nik Vaessen for sharing the N-Best evaluation dataset. This work used the Dutch national e-infrastructure with the support of the SURF Cooperative using grant no. EINF-8324.

\bibliographystyle{IEEEtran}
\bibliography{mybib}

\end{document}